\documentclass[preprint,12pt]{elsarticle}
\usepackage{booktabs}

\usepackage{amssymb}
\usepackage{amsmath}
\usepackage{hyperref} 
\usepackage{etex}
\usepackage[utf8]{inputenc}
\usepackage[T5]{fontenc}

\usepackage{xcolor}

\journal{arXiv}

\begin{document}

\begin{frontmatter}



\title{WriteViT: Handwritten Text Generation with \\ Vision Transformer} 

%

\author[UIT,VNU]{Dang Hoai Nam} 
\author[UIT,VNU]{Huynh Tong Dang Khoa}
\author[UIT,VNU]{Vo Nguyen Le Duy \corref{cor1}}

\affiliation[UIT]{organization={University of Information Technology},
            city={Ho Chi Minh City},
            country={Vietnam}}
            
\affiliation[VNU]{organization={Vietnam National University},
            city={Ho Chi Minh City},
            country={Vietnam}}

            
\cortext[cor1]{Corresponding author. E-mail: duyvnl@uit.edu.vn}      

\begin{abstract}
Humans can quickly generalize handwriting styles from a single example by intuitively separating content from style. Machines, however, struggle with this task, especially in low-data settings, often missing subtle spatial and stylistic cues. Motivated by this gap, we introduce WriteViT, a one-shot handwritten text synthesis framework that incorporates Vision Transformers (ViT), a family of models that have shown strong performance across various computer vision tasks. WriteViT integrates a ViT-based Writer Identifier for extracting style embeddings, a multi-scale generator built with Transformer encoder-decoder blocks enhanced by conditional positional encoding (CPE), and a lightweight ViT-based recognizer. While previous methods typically rely on CNNs or CRNNs, our design leverages transformers in key components to better capture both fine-grained stroke details and higher-level style information. Although handwritten text synthesis has been widely explored, its application to Vietnamese---a language rich in diacritics and complex typography---remains limited. Experiments on Vietnamese and English datasets demonstrate that WriteViT produces high-quality, style-consistent handwriting while maintaining strong recognition performance in low-resource scenarios. These results highlight the promise of transformer-based designs for multilingual handwriting generation and efficient style adaptation. 

\end{abstract}


%


\end{frontmatter}



\section{Introduction}

Despite significant technological advancements, handwritten text continues to play a critical role in various domains, including historical archiving, form processing, and educational assessment. Consequently, handwritten text recognition (HTR) remains a key area of research in document analysis. However, the task poses persistent challenges due to the inherent variability of handwriting. Factors such as diverse individual writing styles, language-specific idiosyncrasies, inconsistent image quality, and uneven illumination conditions complicate the development of robust recognition systems \cite{HANDS-VNOnDB, CVLDataBase, ICFHR2018, NabucoLatin}.

While deep learning---particularly Transformer-based architectures---has shown great promise in improving HTR accuracy, these models generally require large, high-quality labeled datasets to perform effectively. The creation of such datasets, however, is labor-intensive, costly, and often impractical, particularly for low-resource languages or scripts \cite{CVLDataBase}. This bottleneck has spurred interest in handwriting synthesis (HS), which aims to generate realistic handwritten text samples to augment existing datasets or support training in data-scarce environments \cite{RNNSequence, HandwritingT}.

Handwriting synthesis seeks to emulate human writing behavior by producing plausible and stylistically consistent text images for arbitrary content. The goal is not only to capture the lexical correctness of generated text but also to preserve essential stylistic features such as slant, stroke variation, and character spacing. While humans can effortlessly mimic handwriting styles after viewing a few samples, achieving similar generalization in machine learning models remains an open challenge. HS systems must deal with additional complexities, including generating variable-length outputs, handling out-of-vocabulary words, and capturing fine-grained stylistic nuances—challenges that are particularly pronounced in languages like Vietnamese, where the use of tonal diacritics significantly increases orthographic complexity and the availability of training data remains limited.

To address these multifaceted challenges, we introduce WriteViT, a novel handwriting synthesis framework that combines Vision Transformer (ViT) with Generative Adversarial Networks (GANs). Our method is designed to generate high-quality handwritten images that not only match the input text but also adhere closely to the visual style of provided exemplars. By leveraging the ViT's capacity for capturing both local texture details and global structural dependencies, ViT-GAN excels at producing clear, stylistically faithful handwriting from sparse input samples.

The main contributions of this paper are summarized as the following:

\begin{itemize}
    \item We develop a novel approach to replicate handwriting styles using Vision Transformer (ViT) and Generative adversarial networks (GANs), with experiments demonstrating superior accuracy over existing techniques and enhanced support for HTR tasks.
    \item We evaluate WriteViT on the HANDS-VNOnDB~\cite{HANDS-VNOnDB} Vietnamese dataset and the IAM dataset~\cite{IAMDataset}, where it consistently generates handwriting samples with high textual clarity and strong stylistic fidelity, outperforming existing approaches.
    \item We pioneer the application of advanced synthesis to Vietnamese, addressing its unique diacritic challenges and paving the way for broader language diversity in handwriting generation.
    \item Our implementation is available at: 
    \begin{center}
    	\href{https://github.com/hnam-1765/WriteViT}{https://github.com/hnam-1765/WriteViT}
    \end{center}
\end{itemize}

\section{Related Work}
\label{sec2}

Over the past decade, handwriting synthesis has evolved significantly, transitioning from early template-based techniques to advanced deep learning methods. This section surveys pivotal developments in the field, categorized by the architectural paradigms that have propelled the generation of diverse and authentic handwritten text.

\subsection{Early Methods with Recurrent Neural Networks}
\label{subsec21}

Initial efforts in handwriting synthesis focused on online generation using Recurrent Neural Networks (RNNs). A seminal work by Graves~\cite{graves2014generatingsequencesrecurrentneural} introduced Long Short-Term Memory (LSTM) networks~\cite{LSTM} to model text-conditioned, real-valued sequences, predicting stroke points with Gaussian mixture models to synthesize English handwriting trajectories. Building on this foundation, Ha and Eck~\cite{SketchRNN} developed SketchRNN, extending RNN-based synthesis to hand-drawn sketches. More recently, Kotani et al.~\cite{Decoupled} advanced this approach with a decoupled style descriptor model, separating character- and writer-specific styles to produce more lifelike trajectories.

While these RNN-based methods marked significant progress, they face inherent drawbacks. Their sequential processing struggles to capture long-range dependencies in lengthy sequences and limits computational efficiency, as parallelization remains challenging. Additionally, acquiring large-scale trajectory data is impractical, requiring specialized tools like stylus pens and touch screens. In contrast, handwriting images—readily obtainable via cameras and scanners—offer a more feasible basis for synthesis, prompting a shift toward image-based techniques.

\subsection{Generative Approaches}
\label{subsec22}

Recent advances have pivoted toward offline handwriting synthesis, leveraging generative models to overcome earlier limitations~\cite{GANwriting, SmartPatch, ScrabbleGAN}. Kang et al.~\cite{GANwriting} introduced GANwriting, a conditional Generative Adversarial Network (GAN)~\cite{goodfellow2014generative} that combines writer-specific image samples with text conditioning to generate realistic word-level handwritten text. This approach was later extended to sentence-level synthesis~\cite{Kang_2022}, enabling flexible generation of novel text sequences across diverse styles. However, early iterations produced unrealistic pen-level artifacts, a flaw mitigated in SmartPatch~\cite{SmartPatch} through the addition of a patch-level discriminator loss to refine output quality.

In parallel, Fogel et al.~\cite{ScrabbleGAN} proposed ScrabbleGAN, a semi-supervised GAN capable of generating full handwritten sentences with varied styles and content. Bhunia et al.~\cite{HandwritingT} employed an encoder-decoder transformer architecture, harnessing self-attention~\cite{transformer-2017} to capture both short- and long-range contextual dependencies alongside style-content relationships, enhancing offline synthesis robustness. More recent innovations include HTG-GAN~\cite{HTG-GAN}, which disentangles style and content via a dedicated style encoder, and HiGAN~\cite{HiGAN}, which supports both random style generation and imitation from reference images, further expanding the capabilities of generative handwriting synthesis.

\section{Proposed Approach}\label{sec3}

\subsection{Motivation}\label{subsec31}
Recent advances such as Handwriting Transformer (HWT) \cite{HandwritingT} have shown the effectiveness of Transformer architectures in handwriting synthesis by modeling content–style relationships at the character level. However, these models rely on standard Transformers applied directly to image features, which limits their ability to capture spatial structure. In contrast, ViT \cite{vit-2021} processes images as patch sequences, enabling better modeling of both local and global visual context.

Motivated by this, we propose a ViT-based handwriting synthesis framework that replaces the traditional CNNs and CRNN \cite{CRNN} components with unified transformer-based modules. Our model integrates ViT into the Writer Identifier, Generator, and Recognizer, and incorporates multi-scale structure through Conditional Positional Encoding (CPE)~\cite{Conditional-pe-2023} to enhance spatial awareness. Unlike prior work focused mainly on English, we extend our approach to Vietnamese, addressing the underexplored challenge of synthesizing handwriting with rich diacritics and complex compositions from just a single example.

\subsection{Problem Formulation}
\label{subsec32}

In this study, we address the few-shot offline Handwritten Text Generation (HTG) task with a focus on personalized style transfer. Specifically, we assume access to a small set of handwritten word images \( X_i^s = \{x_{ij}\}_{j=1}^P \) authored by a target writer \( i \in W \), with \( P = 15 \) samples as per common practice in prior work (for one-shot variant, we set P = 1). Alongside this, we define a query set \( A = \{a_k\}_{k=1}^Q \), where each \( a_k \) is a word of arbitrary length composed from a comprehensive character set. The objective is to generate a set of synthetic images \( X_i^t \) that render each word in \( A \) in the unique calligraphic style captured by \( X_i^s \).

\subsection{Overall Architecture}

\begin{figure}[!t]
    \centering
    \includegraphics[width=0.99\linewidth]{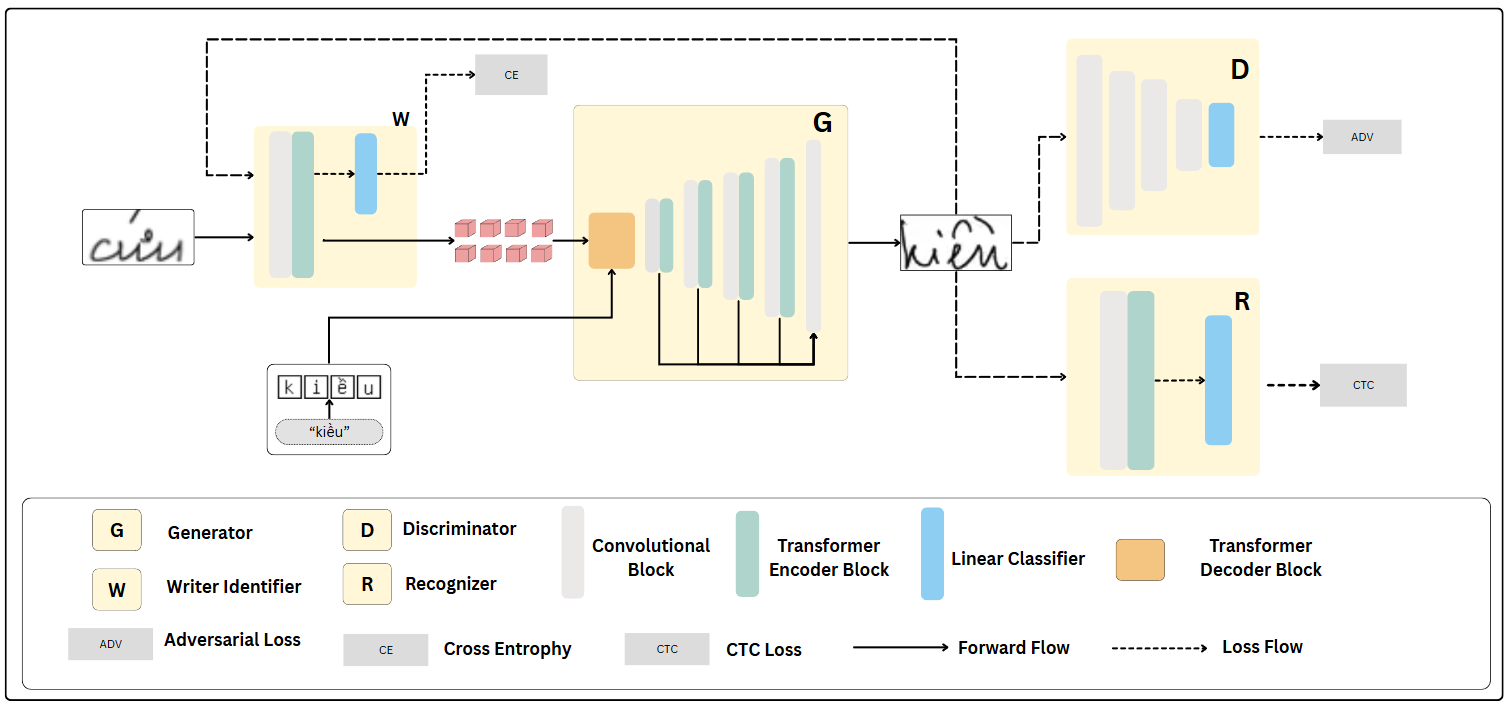}
    \caption{Overview of the proposed WriteViT architecture.}
    \label{WriteViT_architecture}
\end{figure}

Our proposed model, illustrated in Figure~\ref{WriteViT_architecture}, consists of four main components: a Generator ($\mathcal{G}$), a Discriminator ($\mathcal{D}$) a Recognizer ($\mathcal{R}$), and a Writer Identifier ($\mathcal{W}$). Each module is designed with a specific purpose to facilitate high-quality handwriting synthesis and improve downstream handwriting text recognition (HTR). In the following sections, we describe each component in detail.

\label{subsec33}

\subsubsection{Generator}
\label{Generator}

The Generator module ($\mathcal{G}$) in our architecture builds upon the foundation of the Handwriting Transformer (HWT), which models handwriting synthesis using a transformer-based encoder-decoder framework. While HWT captures both global and local writing characteristics effectively, it relies on a single-scale generation pathway, which limits its ability to capture fine-grained variations across different resolutions.

To overcome this limitation, we introduce a multi-scale generation strategy by incorporating a stack of Transformer encoder blocks, each responsible for synthesizing intermediate representations at different spatial resolutions. To enhance the interaction between these representations, we adopt Conditional Positional Encoding using CPE, which allows the model to encode spatial context dynamically across resolutions. This design facilitates richer feature alignment and smoother transitions between coarse and fine details in the generated handwriting.

In the character encoding path, we follow the rendering-based approach proposed in VATr, where each character in the input string is converted into a 16×16 grayscale binary image using a GNU Unifont, a Unicode-complete bitmap font that supports a wide range of scripts and diacritics. These images are flattened and passed through a linear projection layer to obtain fixed-dimensional embeddings. To retain the sequential structure of the input, we add sinusoidal positional encodings, as introduced in \cite{transformer-2017}. The resulting character embeddings serve as keys in the decoder attention mechanism, while the writer-style features (extracted from the Writer Identifier module) are used as queries and values. This asymmetrical key-query pairing enables the model to align character identity with stylistic attributes effectively.

The decoded representations are subsequently processed through multiple Transformer encoder layers, each tailored to a specific scale. These multi-scale outputs are finally aggregated and passed through a lightweight CNN-based decoder to generate the final handwriting image. Compared to HWT, our model produces images at a lower resolution, reducing computational cost while maintaining style consistency and legibility, thanks to the multi-scale design and enhanced positional encoding.

This generator design improves the ability to model stroke-level variations, local distortions, and fine stylistic cues that are critical for producing convincing handwriting images, especially in scenarios with diverse writing styles and low-resolution constraints.

\subsubsection{Discriminator}
\label{Discriminator}

The Discriminator module ($\mathcal{D}$) is tasked with distinguishing between real and generated handwriting images, serving as a critical component in the adversarial learning framework. In our design, we retain the original architecture from the Handwriting Transformer (HWT) model, as it has proven effective in providing stable and meaningful gradient feedback to the Generator. This stability plays a crucial role in guiding the Generator towards producing more visually realistic and stylistically consistent outputs.

\subsubsection{Recognizer}
\label{Recognizer}

The Recognizer module ($\mathcal{R}$) is responsible for transcribing the generated handwriting images into character sequences, serving as a crucial component for enforcing content fidelity. In contrast to most prior works that rely on CRNN-based architectures—typically combining convolutional and recurrent layers—we adopt a more modern approach by replacing the sequence modeling backbone with a Vision Transformer (ViT). This design provides enhanced parallelism, a broader receptive field, and stronger global context modeling.

To enrich the positional encoding within the transformer, we utilize CPE, which allows the model to encode spatial position information based on visual content rather than fixed sinusoidal patterns. The handwriting image is first passed through a lightweight ResNet-18 \cite{resnet18}, which we modify by reducing filter dimensions and removing redundant residual blocks to reduce computational cost. The resulting feature maps are flattened and fed into the ViT for sequence-level feature modeling. The final output consists of contextualized embeddings that represent the transcribed character sequence. This design improves recognition performance while maintaining efficiency.

\subsubsection{Writer Identifier}
\label{Writer Identifie}
The Writer Identifier module ($\mathcal{W}$) serves a dual purpose in our architecture: (i) extracting style embeddings to condition the Generator $\mathcal{G}$, and (ii) classifying writer identities to preserve authorship consistency. In contrast to previous works, which typically rely on CNN-based encoders for this module, we replace the backbone with a pure Vision Transformer (ViT) architecture. This shift enables $\mathcal{W}$ to model long-range dependencies and capture more expressive stylistic representations than conventional approaches.

Given a real handwriting sample, $\mathcal{W}$ encodes the image into a compact embedding vector that encapsulates both local stroke patterns and global stylistic cues. This embedding is passed to the Generator to guide the synthesis process, ensuring the generated handwriting reflects the source writer’s style. Simultaneously, the same embedding is used to predict writer identity, allowing the model to learn features that are both descriptive and discriminative.

To prevent $\mathcal{W}$ from overfitting to generator artifacts, we adopt an alternating training strategy. In the generation phase, $\mathcal{W}$ is used to extract style features but remains frozen—only $\mathcal{G}$ is updated based on reconstruction or adversarial losses. In a separate phase, $\mathcal{W}$ is trained using only real handwriting samples for the classification task, alongside updates to the Discriminator $\mathcal{D}$ and Recognizer $\mathcal{R}$. This staged update scheme stabilizes training, preserves identity fidelity, and ensures $\mathcal{W}$ remains grounded in authentic style representations.

\subsection{Objective Functions}

Our training strategy follows the conventional GAN framework, where the Discriminator $\mathcal{D}$ is optimized to distinguish between real handwriting samples and those synthesized by the Generator $\mathcal{G}$. This adversarial setup drives $\mathcal{G}$ to generate increasingly realistic outputs. We adopt the commonly used hinge loss formulation for adversarial training \cite{HingeLoss}:
\begin{equation}
   \mathcal{L}_{\text{adv}} = \mathbb{E} \left[ \max (1 - \mathcal{D}(X_i^s), 0) \right] + \mathbb{E} \left[ \max (1 + \mathcal{D}(\mathcal{G}(X_i^s, A)), 0) \right] 
\end{equation}
The Recognizer $\mathcal{R}$ is responsible for making sure the generated handwriting accurately reflects the target textual content. This ensures that the generator $\mathcal{G}$ not only replicates the desired style but also reproduce correct textual content accurately. The Recognizer $\mathcal{R}$ is trained using the real images \( X_i^s \) and their labels, and its loss computed on generated samples is backpropagated through $\mathcal{G}$, guiding it to produce textually faithful images. The loss of the Recognizer $\mathcal{R}$ is defined as:
\begin{equation}
\mathcal{L}_R = \mathbb{E}_x \left[ - \sum \log\left( p(y_x \mid \mathcal{R}(x)) \right) \right]
\end{equation}

\noindent Here, \(\mathbf{x} \sim \{ X^s_i, X^t_i \}\), and \( y_x\) is the ground truth transcription string of x.

The writer identifier $\mathcal{W}$ enforces style consistency by preserving the unique calligraphic traits of writer \( i \) from \( X_i^s \). This compels the generator, $\mathcal{G}$, to reproduce the intended style faithfully. The training procedure of $\mathcal{W}$ is similar to the $\mathcal{R}$ module, it is trained using the real images, and its loss on the generated images guides the generator. Specifically, the loss for this module is defined as:
\begin{equation}
\mathcal{L}_W = \mathbb{E}_x \left[ - \sum \log\left( p(i \mid \mathcal{W}(x)) \right) \right]
\end{equation}

Combining the above components, the overall loss used to train our WriteViT model is expressed as:

\begin{equation}
    \mathcal{L} = \mathcal{L}_{adv} + \mathcal{L}_{R} + \mathcal{L}_{W} 
\end{equation}

We also used the technique of stabilizing the training by balancing the gradient magnitudes of $\mathcal{R}$ and $\mathcal{W}$ relative to $\mathcal{D}$ improves performance following the approach in \cite{AdvOnSequence19} similar to HWT \cite{HandwritingT}, we normalize the gradients of $\mathcal{R}$ and $\mathcal{W}$ to match the standard deviation of the discriminator's gradients:

\begin{equation}
\nabla \mathcal{R} \leftarrow \alpha \left( \frac{\sigma_D}{\sigma_R} \cdot \nabla \mathcal{R} \right),
\quad
\nabla \mathcal{W} \leftarrow \beta \left( \frac{\sigma_D}{\sigma_S} \cdot \nabla \mathcal{W} \right).
\end{equation}

Here \( \alpha = 0.7 \), \( \beta = 0.7 \) are hyper-parameters that are fixed during the training of our model.

\section{Experiments}
\label{sec4}

In our experiments, we resized images to a fixed height of 32 pixels and a width of 16 pixels per character in the image label to ensure fair comparisons with other methods. Our approach used the Adam optimizer with a learning rate of 0.00005, trained on a single A100 GPU.

We conducted experiments using the following datasets:

\textbf{IAM Dataset}: The IAM dataset includes 9,862 text lines with about 62,857 English words, written by 500 unique writers. For fair evaluation, we reserved a subset of 160 writers for testing, training our model on images from the remaining 340 writers, consistent with the setup in HWT~\cite{HandwritingT}.

\textbf{HANDS-VNOnDB}: The HANDS-VNOnDB dataset contains 7,296 handwritten Vietnamese text lines, encompassing over 480,000 strokes and more than 380,000 characters from 200 writers. We followed its predefined split, using 106 writers for training and 34 for testing.

To ensure fairness in comparison, we adopted the following strategy: for models such as HWT and VATr, which publicly provide pre-trained checkpoints trained on the above datasets, we directly used the available pre-trained versions. For other baselines like HiGAN, HiGAN+ and ScrabbleGAN, we re-trained them from scratch using the exact same training data partitions as used in HWT and VATr, ensuring that all models were evaluated under the same conditions.
\subsection{Image Quality Assessment}

To evaluate the visual quality of the generated handwriting, we adopt two standard metrics: Fréchet Inception Distance (FID) ~\cite{FID} and Kernel Inception Distance (KID) ~\cite{KID}. Both metrics assess the similarity between real and generated image distributions by comparing features extracted from a pre-trained Inception network. Lower values reflect greater fidelity and realism. These metrics are widely used in image generation literature, including works such as HiGAN and HWT. Our model achieves competitive FID and KID scores, demonstrating its effectiveness in generating visually coherent and high-quality handwriting samples—an outcome attributed to our transformer-based architecture and training strategy. All handwriting images are resized to a fixed resolution of 32×128 pixels to ensure consistency across the training and evaluation pipelines. For images shorter than this size, we apply right-side zero-padding using white pixels to preserve content alignment. In cases where the original image exceeds the target width, we truncate the excess regions to fit within the fixed size.
\begin{table}[!t]
\centering
\caption{Quantitative comparison of handwriting synthesis quality using FID and KID metrics. Lower values indicate higher similarity to real handwriting.}
\label{FID}
\begin{tabular}{lcc}
\toprule
\textbf{Method} & \textbf{FID} & \textbf{KID} \\
\midrule
HiGAN~\cite{HiGAN}             & 17.086 & 1.18 \\
HiGAN+~\cite{higan+}           & 16.114 & 0.81 \\
HWT~\cite{HandwritingT}        & 13.615 & 0.49  \\
VATr~\cite{Vatr}               & 13.577 & 0.47 \\
\textbf{Ours}                  & \textbf{11.102} & \textbf{0.37} \\
\bottomrule
\end{tabular}
\end{table}

As shown in Table ~\ref{FID} , our model achieves the lowest FID (11.102) and KID (0.37) scores among all compared methods. This indicates a significantly closer match to the real handwriting distribution, confirming that our ViT-based architecture with multi-scale generation and enhanced style encoding yields more realistic and visually coherent handwriting. These results highlight the effectiveness of our method in producing high-fidelity handwritten images, even under limited data conditions.

\subsection{Fine-Grained Evaluation by Vocabulary and Style}

To further evaluate the quality of synthetic handwriting under more realistic and diverse scenarios, we conduct a fine-grained analysis of FID scores across different lexical and stylistic conditions. Specifically, we consider four distinct evaluation settings: 
\begin{itemize}
    \item \textbf{In-Vocabulary words and Seen style(IV-S)}: Words present in the training vocabulary, written in handwriting styles seen during training.
    \item \textbf{In-Vocabulary words and  Unseen style(IV-U)}: Words in vocabulary, but written in previously unseen handwriting styles.
    \item \textbf{Out-of-Vocabulary words and Seen style(OOV-S)}: Words not seen during training, but written in seen styles.
     \item \textbf{Out-of-Vocabulary words and UnSeen style(OOV-U)}: Words and styles both unseen during training.    
\end{itemize}

For each condition, we generate 25,000 images using the respective synthesis models, and compute the Fréchet Inception Distance (FID) between generated and real samples. This setup allows us to assess a model's ability to generalize across lexical novelty (OOV) and stylistic variation (Unseen writers).
\begin{table}[!t]
\centering
\caption{FID scores under different evaluation conditions: In/Out-of-Vocabulary and Seen/Unseen styles (lower is better). Each condition includes 25,000 generated samples.}
\label{I-OOV}
\begin{tabular}{lcccc}
\toprule
\textbf{Method} & \textbf{IV-S} & \textbf{IV-U} & \textbf{OOV-S} & \textbf{OOV-U} \\
\midrule
HiGAN~\cite{HiGAN}             & 36.02 & 40.94& 36.54 & 41.35  \\
HiGAN+~\cite{higan+}           & 34.90 & 37.78 & 34.66 & 37.75 \\
HWT~\cite{HandwritingT}        & 26.46 & 29.86 & \textbf{26.47} & 29.68 \\
VATr~\cite{Vatr}               & 26.73 & 29.94 & 26.82 & \textbf{29.50} \\
\textbf{Ours}                  & \textbf{26.26} & \textbf{29.46}  & 27.56 & 30.87 \\
\bottomrule
\end{tabular}
\end{table}

Results reported in Table~\ref{I-OOV} indicate that our model consistently outperforms HiGAN and HiGAN+ across all four evaluation settings and achieves competitive performance compared to HWT and VATr, particularly under the most challenging scenario of out-of-vocabulary words with unseen writing styles (OOV-U). While our FID scores are slightly higher than those of HWT and VATr in some easier settings (e.g., IV-S), our model maintains robust generalization when both content and style are novel. These results highlight the strength of our ViT-based handwriting synthesis framework in handling cross-style and cross-lexical generalization, which is critical for real-world applications involving diverse writing inputs.
\subsection{Improving HTR with Synthetic Data}

This evaluation is central to our study. The primary objective of synthetic handwriting generation is to improve Handwritten Text Recognition (HTR) performance, especially in low-resource settings. To simulate this scenario, we first trained a Transformer-based OCR ~\cite{TrOCR} model using 5,000 real handwriting images from the IAM dataset as the baseline. To evaluate the benefit of synthetic data, we then augmented the training set with 25,000 additional synthetic images generated by each comparative method, including ours. The combined dataset (5000 real images + 25000 synthetic images) was used to retrain the HTR model, and performance was evaluated using standard metrics. This protocol allows us to directly assess how well each synthesis model contributes to improving recognition accuracy when real data is limited.

We measured HTR performance using three key metrics:
\begin{itemize}
    \item \textbf{Character Error Rate (CER)} -- The ratio of incorrect characters (substitutions, insertions, deletions) to total characters in the ground truth; lower values reflect higher accuracy.
    \item \textbf{Word Error Rate (WER)} -- The ratio of incorrect words to total words in the ground truth; lower values indicate better word-level recognition.
    \item \textbf{Normalized Edit Distance (NED)} -- The normalized similarity between predicted and ground truth text; lower values signify closer matches.
\end{itemize}
 
\begin{table}[!t]
\centering
\caption{HTR performance (lower is better) using 5,000 real images and synthetic data generated by each method on the IAM dataset.}
\label{HTR}
\begin{tabular}{lccc}
\toprule
\textbf{Method} & \textbf{WER} & \textbf{CER} & \textbf{NED} \\
\midrule
5000 real images & 36.66 & 25.50 & 19.13\\
ScrabbleGAN~\cite{ScrabbleGAN} & 7.56 & 4.06 & 3.52 \\
HiGAN~\cite{HiGAN}       & 6.37 & 3.25 & 3.21 \\
HiGAN+~\cite{higan+}           & 7.25 & 6.07 & 4.28 \\
HWT~\cite{HandwritingT}               & 6.85 & 3.48 & 3.47 \\
VATr~\cite{Vatr}             & 6.21 & 3.14 & 3.09 \\
\textbf{Ours}                  & \textbf{5.76} & \textbf{3.13} & \textbf{3.09} \\
\bottomrule
\end{tabular}
\end{table}

As shown in Table ~\ref{HTR}, our method outperforms all baselines across the three key HTR metrics—WER, CER, and NED on the IAM dataset. These results demonstrate the effectiveness of our ViT-based handwriting synthesis framework in generating high-quality, stylistically coherent images that enhance recognition performance. The improvements are particularly notable under low-resource conditions, where only 5,000 real images were available. Our model's ability to provide such gains highlights its practical potential for augmenting HTR systems in scenarios with limited annotated data.

\subsection{Ablation Study}
To assess the contribution of individual components in our proposed architecture, we conducted an ablation study on the IAM dataset. Beginning with a baseline model composed of standard CNN and CRNN components, we incrementally integrated key architectural enhancements and evaluated their effect on synthesis quality using Fréchet Inception Distance (FID) as the primary metric.
\begin{table}[!t]
\centering
\caption{Ablation study on key components of our model evaluated on IAM dataset.}
\label{Ablation Study}
\begin{tabular}{lc}
\toprule
\textbf{Method} & \textbf{FID} \\
\midrule
 
Base (CNN + CRNN)       & 13.615 \\
+ ViT at Generator (one-shot)             & 13.19\\
+ Multi-scale Transformer Encoder Blocks         &12.322\\
+ ViT for Recognizer and Writer Identifier & \textbf{11.102}\\

\bottomrule
\end{tabular}
\end{table}

Table ~\ref{Ablation Study} presents the results. Introducing ViT at the Generator with one-shot conditioning yields a clear reduction in FID, suggesting that early integration of transformer-based modeling improves the visual realism of the synthesized handwriting. Further incorporating multi-scale Transformer encoder blocks enhances spatial diversity and leads to an even lower FID score. Finally, replacing both the Recognizer and Writer Identifier modules with ViT-based architectures results in the lowest FID, indicating the model's ability to generate highly realistic and coherent handwriting images.

These findings highlight how each architectural modification contributes differently to synthesis performance. While ViT-based generation offers immediate quality gains, the full ViT-based model with multi-scale design achieves the best overall realism. This supports the value of a modular design in synthetic handwriting systems, where components can be flexibly adapted based on the target application. Depending on whether the goal is image realism for data augmentation or recognition alignment, users can selectively enable or simplify architectural components (e.g., multi-scale, ViT recognizer, writer encoder) to meet specific needs.

\subsection{Model Size}

\begin{table}[!t]
\centering
\caption{Comparison of existing models and our proposed model in terms of size in megabytes. Only the modules necessary to store to perform generation, that is generator (Gen) and encoder (Enc), are considered.}
\label{size}
\begin{tabular}{lccc}
\toprule
\textbf{Method} & \multicolumn{3}{c}{\textbf{Size (MB)}} \\
\cmidrule(lr){2-4}
 & \textbf{Gen} & \textbf{Enc} & \textbf{Total} \\
\midrule
ScrabbleGAN~\cite{ScrabbleGAN} & 81.8 & N/A & 81.8 \\
HWT~\cite{HandwritingT}        & 80.7 & 50.6 & 131.3 \\
HiGAN~\cite{HiGAN}             & 38.6 & 20.5 & 59.1 \\
HiGAN+~\cite{higan+}           & \textbf{15.0} & \textbf{6.7}  & \textbf{21.7} \\
\textbf{Ours}                  & 32.2 & 10.4 & 42.6 \\
\bottomrule
\end{tabular}
\end{table}

The table~\ref{size} highlights the compactness of our proposed model in comparison with existing handwriting synthesis methods. Notably, our architecture achieves a total size of only 42.6 MB, significantly smaller than the original HWT model (131.3 MB) on which it is conceptually based. Among all methods compared, our model ranks among the most lightweight, while still maintaining high-quality generation and recognition performance.

This compact size emphasizes the model’s suitability for real-world deployment, especially on resource-constrained platforms such as mobile devices and embedded systems. It is worth noting that this comparison only includes the modules essential for image generation: the Generator and Encoder. Excluding auxiliary components ensures a fair and practical evaluation aligned with real usage scenarios in handwriting image synthesis applications.

\subsection{Visual Comparison}
\subsubsection{Generation}
\begin{figure}[!t]
\centering
\includegraphics[width=\linewidth]{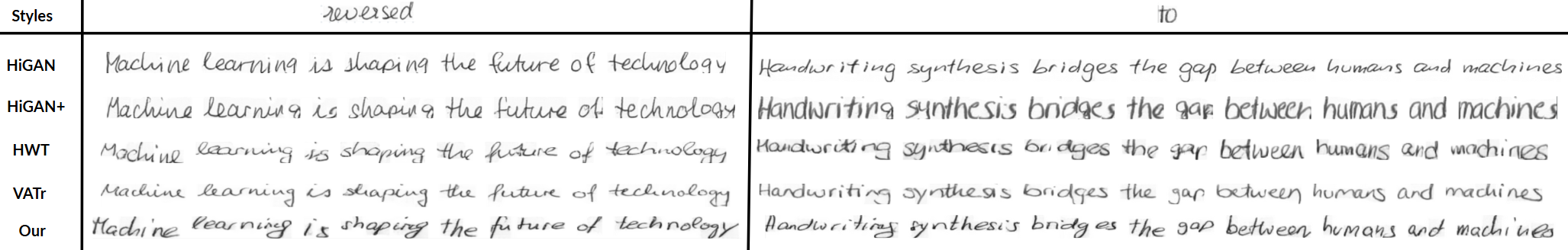}
\caption{Qualitative comparison of generated handwriting from different models given the same style and content inputs. Each row represents a different model, while each column corresponds to a specific input sentence.}
\label{fig:qualitative_gen}
\end{figure}
Figure~\ref{fig:qualitative_gen} illustrates qualitative comparisons between our model and several existing handwriting synthesis methods, including HiGAN, HiGAN+, HWT, and VATr. Each model was conditioned on the same style reference and content input across all rows. Our model demonstrates visually consistent and stylistically faithful handwriting generation. Compared to HiGAN and HiGAN+, our samples exhibit better stroke continuity, spacing alignment, and less visual distortion. While HWT and VATr also perform well, our method maintains greater coherence in difficult regions, such as complex letter transitions (e.g., "bridges the gap", "technology"). This suggests improved spatial encoding and content-style disentanglement, particularly under challenging content inputs.

\subsubsection{Reconstruction}

\begin{figure}[!t]
\centering
\includegraphics[width=\linewidth]{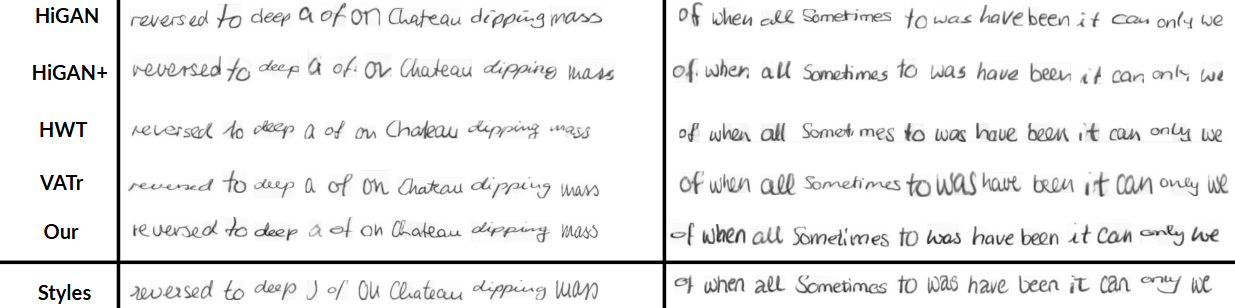}
\caption{Qualitative reconstruction results. Each row corresponds to a different model, conditioned on the same ground-truth style (bottom row) and target text. The goal is to reproduce the target content while preserving the handwriting style.}
\label{fig:reconstruction}
\end{figure}

Figure~\ref{fig:reconstruction} shows the reconstruction results for various models, where each is given a ground-truth style image and target content, and is expected to reproduce the handwriting as closely as possible. Our model achieves strong content preservation while maintaining a high degree of style fidelity. Compared to HiGAN and HiGAN+, our reconstructions exhibit smoother strokes, more consistent character sizing, and better spacing alignment. While HWT and VATr also retain style features effectively, our model more accurately replicates subtle style cues such as slant, baseline flow, and stroke sharpness, as evidenced in sequences like “Chateau dipping mass” and “to was have been”. The qualitative results suggest that our ViT-based architecture is well-suited for handwriting reconstruction tasks, offering a balance between visual realism and fine-grained style adherence.

\subsection{Applications to Other Languages}
\subsubsection{Image Quality Assessment}
\begin{table}[!t]
\centering
\caption{Quantitative comparison of handwriting synthesis quality using FID and KID metrics on VNOnDB dataset.}
\label{FIDVN}
\begin{tabular}{lcc}
\toprule
\textbf{Method} & \textbf{FID} & \textbf{KID} \\
\midrule
ScrabbleGan~\cite{ScrabbleGAN} & 19.2319 & 1.11  \\  
HWT~\cite{HandwritingT}        & 9.8507 & 0.72  \\
VATr~\cite{Vatr}               & 23.8826 & 2.72 \\
HiGAN~\cite{HiGAN}             & 11.2572 & 0.79 \\
\textbf{Ours}                  & \textbf{6.1785} & \textbf{0.51} \\
\bottomrule
\end{tabular}
\end{table}

Table ~\ref{FIDVN}  presents a quantitative comparison of handwriting synthesis quality on the VNOnDB dataset, using FID and KID as evaluation metrics. Among all methods, our model achieves the lowest FID (6.1785) and lowest KID (0.512) scores, indicating a substantially closer alignment with the real handwriting distribution. These improvements suggest that our architecture is particularly well-suited for Vietnamese handwriting synthesis, which is inherently more challenging due to the presence of complex diacritics and diverse stroke patterns.

Compared to strong baselines such as HWT (FID: 9.8507) and HiGAN (FID: 11.2572), our method shows a marked reduction in distributional divergence, confirming the benefit of integrating ViT modules and multi-scale design. Notably, VATr, despite employing transformer mechanisms, performs poorly in this setting (FID: 23.8826), reinforcing the importance of architectural balance and targeted feature extraction when dealing with low-resource scripts like Vietnamese.

These results validate the effectiveness and generalizability of our proposed framework in multilingual handwriting synthesis, particularly for underrepresented writing systems.
\subsubsection{Visual Comparison}
\begin{figure}[!t]
\centering
\includegraphics[width=0.9\linewidth]{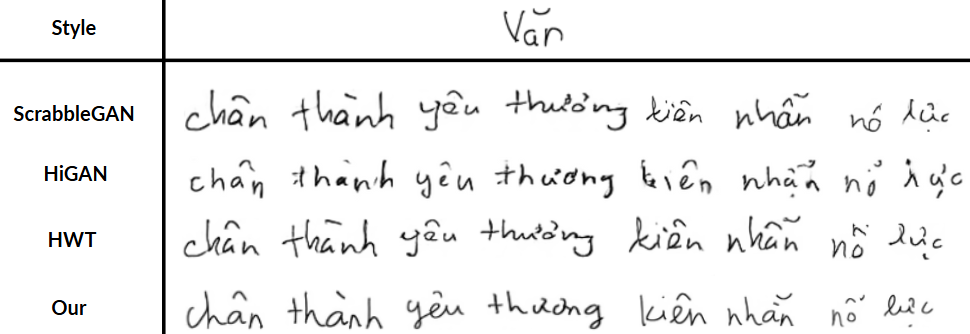}
\caption{Qualitative comparison of Vietnamese handwriting generation on the VNOnDB dataset. Each model is conditioned on the same input text and writer style.}
\label{fig:vnon_generation}
\end{figure}

To evaluate cross-lingual generalization, we conducted experiments on VNOnDB, a Vietnamese handwriting dataset. Figure~\ref{fig:vnon_generation} presents qualitative comparisons across multiple models when tasked with generating a Vietnamese sentence: “chân thành yêu thương kiên nhẫn nỗ lực”. While all models attempt to preserve the writing style, our method more consistently generates complex diacritics (e.g., “ỗ”, “ậ”, “ư”)” with correct placement and spacing. ScrabbleGAN and HiGAN show frequent omissions or distortions in accent marks, while HWT occasionally misplaces tonal diacritics. In contrast, our model preserves diacritic structure and overall legibility, producing clean and well-formed Vietnamese characters. These results demonstrate the robustness of our ViT-based framework when applied to languages with rich diacritic systems, such as Vietnamese, and underscore its potential for multilingual handwriting synthesis tasks.
\section{Conclusion}\label{sec5}

In this paper, we presented a novel handwriting synthesis framework based on Vision Transformers (ViT), WriteViT, designed to improve the generation of realistic and style-consistent handwritten text. Unlike prior approaches that rely heavily on CNN- or RNN-based modules, our method leverages ViT at multiple stages of the architecture, including the Generator, Recognizer, and Writer identifier, along with a multi-scale design to better capture both local and global spatial patterns. Extensive experiments on English (IAM) and Vietnamese (VNOnDB) datasets demonstrate that our model consistently achieves competitive or superior performance in terms of visual quality (FID, KID) and downstream handwriting text recognition (CER, WER, NED), even under challenging cross-style and cross-lingual settings. Qualitative results further highlight the model’s ability to generate stylistically coherent and legible handwriting, including complex scripts with diacritics such as Vietnamese. Our findings indicate that ViT-based synthesis architectures offer strong potential for improving handwritten text generation, particularly in low-resource or multilingual scenarios. Future work could explore extending this approach to more complex sequence modeling tasks or integrating text-to-handwriting personalization for specific writers or domains.

%

\bibliographystyle{elsarticle-num}
\bibliography{references} 

\end{document}